\documentclass[runningheads]{llncs}
\usepackage[T1]{fontenc}
\usepackage{lscape}
\usepackage{graphicx}
\usepackage{amsmath,amssymb,amsfonts}
\usepackage{algorithmic}
\usepackage{graphicx}
\usepackage{textcomp}
\usepackage{xcolor}
\usepackage{float}
\restylefloat{table}
\usepackage{lscape}
 \usepackage[numbers]{natbib}
\usepackage{graphicx}
\usepackage{caption}
\usepackage{subcaption}
\usepackage{booktabs,makecell}
\usepackage{multirow}
\usepackage{pdflscape}
\usepackage{caption}
\usepackage{hyperref}
\usepackage{rotating}
\usepackage[export]{adjustbox}
\usepackage{comment}
\usepackage{amsmath}
\usepackage{xcolor}
\usepackage{tikz}
\usetikzlibrary{shapes,arrows}
\usepackage{placeins}
\usepackage{graphics}
\usepackage{xcolor}
\usepackage{mathrsfs}
\usepackage{blindtext}
\usepackage{hyperref}

\usepackage{algorithm}
\usepackage{algorithmic}

\usepackage{epstopdf}
\def\BibTeX{{\rm B\kern-.05em{\sc i\kern-.025em b}\kern-.08em
    T\kern-.1667em\lower.7ex\hbox{E}\kern-.125emX}}
\begin{document}
\title{Wave-RVFL: A Randomized Neural Network Based on Wave Loss Function}
\author{M. Sajid \and A. Quadir \and
M. Tanveer}
\authorrunning{M. Sajid et al.}
%
\institute{Indian Institute of Technology Indore, Simrol, Indore, India 
\email{\{phd2101241003,mscphd2207141002,mtanveer\}@iiti.ac.in}\\
}
\maketitle              
\begin{abstract}
The random vector functional link (RVFL) network is well-regarded for its strong generalization capabilities in the field of machine learning. However, its inherent dependencies on the square loss function make it susceptible to noise and outliers. Furthermore, the calculation of RVFL's unknown parameters necessitates matrix inversion of the entire training sample, which constrains its scalability. To address these challenges, we propose the Wave-RVFL, an RVFL model incorporating the wave loss function. We formulate and solve the proposed optimization problem of the Wave-RVFL using the adaptive moment estimation (Adam) algorithm in a way that successfully eliminates the requirement for matrix inversion and significantly enhances scalability. The Wave-RVFL exhibits robustness against noise and outliers by preventing over-penalization of deviations, thereby maintaining a balanced approach to managing noise and outliers. The proposed Wave-RVFL model is evaluated on multiple UCI datasets, both with and without the addition of noise and outliers, across various domains and sizes. Empirical results affirm the superior performance and robustness of the Wave-RVFL compared to baseline models, establishing it as a highly effective and scalable classification solution. The source codes and the Supplementary Material are available at \url{https://
github.com/mtanveer1/Wave-RVFL}\\.
\end{abstract}
\section{Introduction}
Artificial neural networks (ANNs) are machine learning models designed to mimic the structure and function of the human brain's neural system. In ANNs, nodes, or ``neurons'', are connected in layers that collaborate to process, analyze, and transmit information, enabling the network to make decisions. ANNs have shown success in various fields, including stock market prediction \cite{lu2024trnn}, diagnosis of Alzheimer's disease \cite{tanveer2024ensemble,10561527,10064117}, significant memory concern \cite{Sajid2024smc}, speech enhancement \cite{jain2024lstmse}, DNA-binding proteins \cite{quadir2024multiview} and among others.

Despite their advantages, ANN models face several challenges, such as slow convergence, difficulties with local minima, and sensitivity to learning rates \cite{sajid2024intuitionistic, 10552388}. To address these issues, randomized neural networks (RNNs) \cite{suganthan2018non}, such as random vector functional link (RVFL) neural networks \cite{pao1994learning}, have been introduced. The RVFL model features a single hidden layer with weights from the input to the hidden layer that are randomly generated and fixed during training. It also includes direct connections from the input layer to the output layer, which serve as a built-in regularization mechanism, enhancing the RVFL's generalization ability \cite{zhang2016comprehensive}. The training process focuses solely on determining the weights of the output layer, which can be efficiently accomplished using closed-form or iterative methods. 

Several enhanced versions of the standard RVFL model have been developed to improve its generalization performance, making it more robust and effective for practical applications \cite{malik2022random}. The traditional RVFL model assigns equal weight to each sample, which makes it susceptible to noise and outliers. To address this issue, a modified version called intuitionistic fuzzy RVFL (IFRVFL) was proposed in \cite{malik2022alzheimer}, which uses fuzzy membership and nonmembership functions to assign an intuitionistic fuzzy score to each sample. In the standard RVFL, original features are transformed into randomized features, leading to potential instability. To counter this, \citet{zhang2019unsupervised} incorporated a sparse autoencoder with \( l1 \)-norm regularization into the RVFL, resulting in the SP-RVFL model. This model mitigates instability caused by randomization and enhances the learning of network parameters compared to the traditional RVFL. \citet{li2021discriminative} proposed another variant, the discriminative manifold RVFL (DMRVFL), which employs manifold learning techniques. DMRVFL replaces the rigid one-hot label matrix with a flexible soft label matrix to better utilize intraclass discriminative information and increase the distance between interclass samples. Recently, In \cite{MalikGraph2022}, authors developed the graph embedded intuitionistic fuzzy weighted RVFL (GE-IFWRVFL), which integrates subspace learning criteria within a graph embedding framework. This approach aims to improve the RVFL's performance by leveraging the graphical information of the dataset into the learning process. Additionally, to address class imbalance issues associated with RVFL, \citet{ganaie2024graph} proposed the graph embedded intuitionistic fuzzy RVFL for class imbalance learning (GE-IFRVFL-CIL). This model incorporates a weighting scheme based on the class imbalance ratio to manage class imbalance effectively. To incorporate human-like decision-making capabilities, \citet{sajid2024neuro} introduced the neuro-fuzzy RVFL, which operates on the IF-THEN logic principle, thereby enhancing the interpretability of the RVFL model. Further, an efficient angle-based twin RVFL (ATRVFL) is proposed in \cite{mishra2024efficient} to effectively tackle binary classification problems.

Despite recent advancements in RVFL, its performance is heavily influenced by the training samples' targets. The RVFL's reliance on the square loss function \cite{wang2020comprehensive} assumes that errors derived from training targets follow a Gaussian distribution \cite{wang2020robust}. However, real-world data often contain noise and outliers, violating this assumption. Outliers with large deviations are disproportionately considered during training, reducing RVFL's robustness to outliers. Additionally, by equally weighting all errors, RVFL with square error loss can over-penalize small errors and excessively punish larger ones, skewing its learning process. The square error's non-robustness to noise further limits its effectiveness. Optimization challenges from steep gradients caused by large errors can also lead to numerical instability. Collectively, these drawbacks hinder the RVFL's performance with square error loss, particularly in noisy or outlier-prone datasets. Next, calculating the parameters of RVFL involves the computation of the matrix inverse of the whole training matrix, which may be intractable in large-scale problems. 

Recently, the wave loss function, introduced by \cite{AKHTAR2024110637}, has been employed in support vector machines (SVM) with some desired claims. It is said to have a trend towards smoothness, insensitivity to noise, and resilience against outliers. This raises an intriguing question: can the wave loss function be effectively utilized in the optimization function of the RVFL network? If so, can we formulate a solution to this optimization problem, and will the integrated model (RVFL with wave loss function) exhibit the same robustness against noise and outliers?

Inspired by these promising properties of the wave loss function and motivated by the potential answers to these questions, we propose the Wave-RVFL model, an RVFL model based on the wave loss function. In developing Wave-RVFL, we aimed to preserve the inherent optimization formulation of the RVFL as much as possible, simply replacing the square loss function with the wave loss function. The proposed optimization problem of Wave-RVFL is solved using the adaptive moment estimation (Adam) algorithm \cite{kingma2014adam}. 

In the following sections, we demonstrate that the proposed Wave-RVFL model exhibits superior generalization performance, showcasing remarkable robustness against noise and outliers compared to baseline models. Additionally, we bypass the need for matrix inversion when calculating the parameters of the RVFL, significantly enhancing its scalability. This simple yet effective approach strengthens the RVFL’s resilience against noise and outliers and makes it more practical for large-scale applications.
The paper's key highlights are as follows:
\begin{enumerate}
\item We propose the Wave-RVFL model, integrating the wave loss function into RVFL while applying penalties for classifier construction.
\item To solve the inherent optimization problem of Wave-RVFL, we employ the Adam algorithm, chosen for its efficiency in managing memory and its efficacy in addressing large-scale problems; this marks Adam's first application in solving an RVFL problem.
\item Our formulation of the Wave-RVFL optimization problem is designed to bypass the need for matrix inversion when calculating parameters. This approach significantly enhances the scalability of the Wave-RVFL.
\item The proposed Wave-RVFL demonstrates robustness against noise and outliers by preventing excessive penalization of deviations, thereby ensuring a balanced treatment of noise and outliers.
\item We evaluate the performance of the proposed Wave-RVFL model using benchmark UCI datasets, which vary in domain and size. These datasets are tested with and without added noise and outliers, allowing us to compare the performance of Wave-RVFL against existing models.
\end{enumerate}
The succeeding sections of this paper are structured as follows: Section \ref{Related_works} introduces square error loss and RVFL. Section \ref{proposed_work} details the mathematical framework and optimization algorithm of the proposed Wave-RVFL model. Experimental results, analyses and discussions of proposed and existing models are discussed in Section \ref{experiments}. Section \ref{conclusions} outlines the conclusion and future research directions.
\section{Related Work}
\label{Related_works}
In this section, we fix some notations and then discuss the RVFL model. The square loss function is discussed in Section S.I of the supplementary materials. 
\vspace{-2mm}
\subsection{Notations}
Let the training dataset be denoted as $\mathcal{X} = \{(x_i, y_i) \mid i \in \{1, 2, \ldots, n\}\}$, where $x_i \in \mathbb{R}^{1 \times m}$ represents the input features and $y_i \in \{+1, -1\}$ represents the corresponding target vector, with $n$ being the total number of training samples. Here, $m$ denotes the number of attributes. The transpose operator is represented by $(\cdot)^T$. The matrices of input and output samples are given by $X = [x_1^T, x_2^T, \ldots, x_n^T]^T$ and $Y = [y_1^T, y_2^T, \ldots, y_n^T]^T$, respectively.
\subsection{Random Vector Functional Link (RVFL) Network}
\begin{figure}
    \centering
    \includegraphics[width=0.5\linewidth]{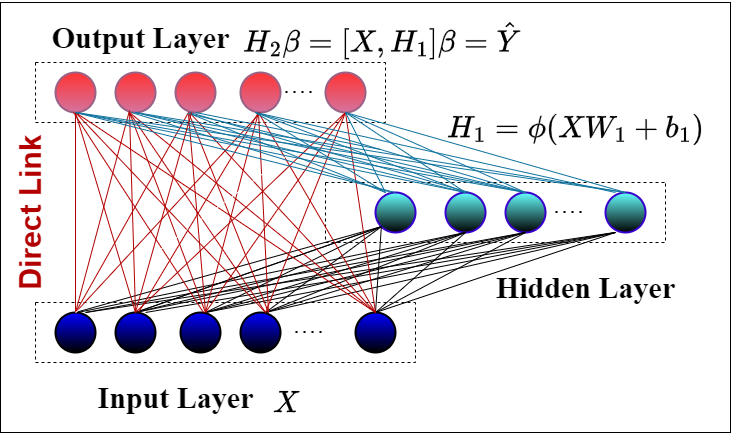}
    \caption{RVFL Neural Network.}
    \label{fig:RVFL}
\end{figure}

The RVFL network, introduced by \citet{pao1994learning}, is a type of single-layer feed-forward neural network that includes three layers: the input layer, the hidden layer, and the output layer. In the RVFL network, the connections between the input and hidden layers, as well as the hidden layer biases, are randomly set at the start and remain unchanged during training. The input samples' original features are directly linked to the output layer. The output layer weights are calculated using the least squares method or the Moore-Penrose inverse. Figure \ref{fig:RVFL} demonstrates the architecture of the RVFL.
The hidden layer matrix (with $N$ number of hidden nodes in the hidden layer), represented as \(H_1\), is defined as follows:
\begin{align}
    H_1  = \phi(XW_1 + b_1) \in \mathbb{R}^{n \times N},
\end{align}
where \(W_1 \in \mathbb{R}^{m \times N}\) represents the weight matrix, initialized randomly with values drawn from a uniform distribution over \([-1, 1]\), \(b_1 \in \mathbb{R}^{n \times N}\) is the bias matrix, and \(\phi\) is the activation function. 
The output layer weights are calculated using the following matrix equation:
\begin{align}
\label{eq:3}
    H_2\beta=\begin{bmatrix}
            X & H_1
        \end{bmatrix}\beta = \hat{Y}.
\end{align}
Here, $H_2=\begin{bmatrix}
            X & H_1
        \end{bmatrix}$, the weight matrix \(\beta \in \mathbb{R}^{(m + N) \times 1}\) connects the combined input and hidden nodes to the output nodes. \(\hat{Y}\) denotes the predicted output. The optimization problem, derived from Eq. (\ref{eq:3}), is formulated as follows:
\begin{align}
\label{eq:4}
    \beta_{min} = \underset{\beta}{\arg\min} \frac{\mathcal{C}}{2}\|H_2\beta - Y\|^2 + \frac{1}{2}\|\beta\|^2.
\end{align}
 In the optimization problem of RVFL, we observe that the square error loss, i.e., \(\|H_2\beta - Y\|^2\), is used to calculate the prediction error.
        
        The optimal solution of Eq. (\ref{eq:4}) can be calculated as follows:
\begin{align}
    (\beta)_{min}=\left\{\begin{array}{ll} 
{H_2}^t\left({H_2} {H_2}^t+\frac{1}{\mathcal{C}} I\right)^{-1} {Y}, & n<(m+N),\vspace{3mm} \\ 
\left({H_2}^t {H_2}+\frac{1}{\mathcal{C}} I\right)^{-1} {H_2}^{t} {Y}, & (m+N) \leq n, \end{array}\right.
\end{align}
where \(\mathcal{C} > 0\) is the regularization parameter, and \(I\) denotes the identity matrix with appropriate dimensions.
\section{Proposed Work}
\label{proposed_work}
In this section, we propose a novel Wave-RVFL model, i.e., the random vector functional link network based on the wave loss function. We further provide the mathematical formulation, its solution using Adam algorithms and the proposed algorithm. Wave-RVFL capitalizes on the asymmetry inherent in the wave loss function to dynamically apply penalties at the instance level for misclassified samples. The proposed Wave-RVFL demonstrates robustness against noise and outliers by preventing excessive penalization of deviations, thereby ensuring a balanced treatment of noise and outliers. 

We start by reviewing some of the properties of the wave loss function and then dive into the formulation of Wave-RVFL.
\subsection{Wave Loss Function}
\begin{figure*}[ht!]
\begin{minipage}{.25\linewidth}
\centering
\subfloat[$\gamma=0.5$]{\label{fig:2a}\includegraphics[scale=0.16]{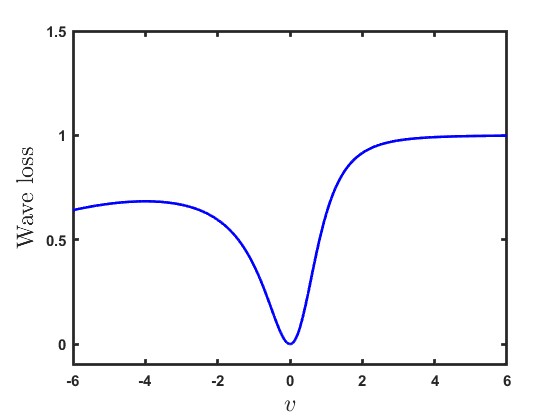}}
\end{minipage}
\begin{minipage}{.25\linewidth}
\centering
\subfloat[$\gamma=1$]{\label{fig:2b}\includegraphics[scale=0.16]{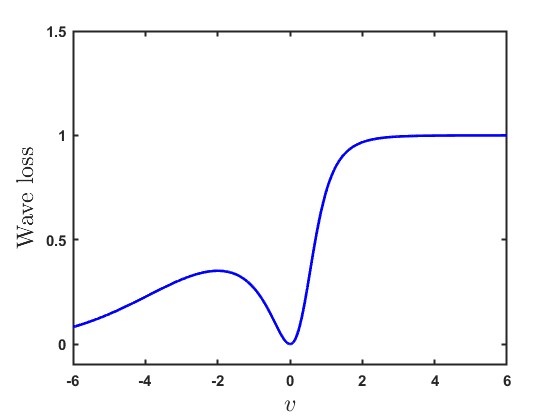}}
\end{minipage}
\begin{minipage}{.24\linewidth}
\centering
\subfloat[$\gamma=1.5$]{\label{fig:2c}\includegraphics[scale=0.16]{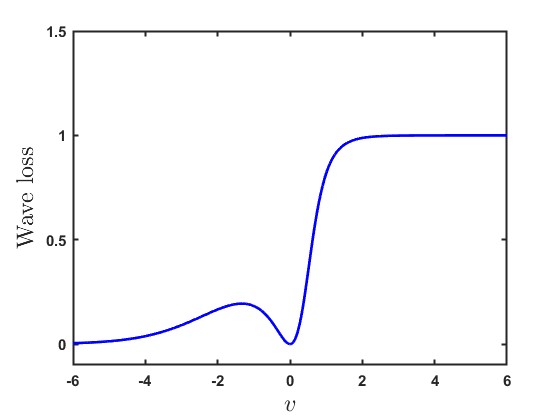}}
\end{minipage}
\begin{minipage}{.24\linewidth}
\centering
\subfloat[$\gamma=2$]{\label{fig:2d}\includegraphics[scale=0.16]{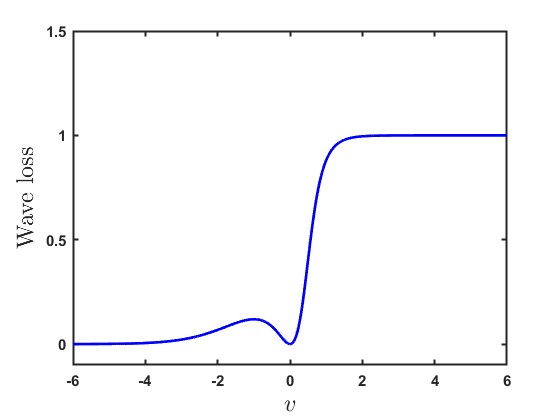}}
\end{minipage}
\caption{A graphical representation of the wave loss function is provided, with \( \eta \) set to $1$, and varying values of \( \gamma \).}
\label{An illustration of the wave loss function}
\end{figure*}

We present the formulation of the wave loss function, engineered to be resilient to outliers, robust against noise, and smooth in its properties, which can be formulated as:
\begin{align}
\label{Wave:1}
    \mathcal{L}_{wave}(v) = \frac{1}{\eta}\left( 1 - \frac{1}{1+ \eta v^2 \exp{(\gamma v)}} \right ), \hspace{0.2cm} \forall \hspace{0.1cm} v \in \mathbb{R},
\end{align}
where \( \eta \in \mathbb{R}^+ \) represents the bounding parameter and \( \gamma \in \mathbb{R} \) denotes the shape parameter. Figure \ref{An illustration of the wave loss function} visually illustrates the wave loss function. The wave loss function exhibits the following properties \cite{AKHTAR2024110637}:
\begin{enumerate}
    \item It is bounded, smooth, and non-convex function.
    \item The wave loss function introduces two essential parameters: the shape parameter \( \gamma \), dictating the shape of the loss function, and the bounding parameter \( \eta \), determining the loss function threshold values. 
    \item The wave loss function is infinitely differentiable and hence continuous.
    \item The wave loss function showcases resilience against outliers and noise insensitivity. With loss bounded to \( \frac{1}{\gamma} \), it handles outliers robustly while assigning loss to samples with \( v \leq 0 \), displaying resilience to noise.
    \item As \( \gamma \) tends to infinity, for a fixed \( \eta \), the wave loss function converges point-wise to the \( 0 - \frac{1}{\eta} \) loss, expressed as:
     \begin{align}
        \mathcal{L}_{0-1}(v) = \left\{
  \begin{array}{lr} 
      0, & \text{if}~ v\leq 0, \\
      \frac{1}{\eta}, & \text{if}~ v > 0. 
      \end{array}
    \right.
    \end{align}
    Furthermore, the wave loss converges to the ``$0 - 1$'' loss when \( \eta = 1 \).
\end{enumerate}
\vspace{-8mm}
\vspace{-10mm}
\subsection{Optimization Problem: Random Vector Functional Link Network with Wave Loss Function (Wave-RVFL)}
In this subsection, we give the mathematical formulation of the proposed Wave-RVFL as follows.
\begin{align}
\label{Eq:1}
    &\min f(\beta) = \min\frac{1}{2}\|\beta\|^2 + \frac{\mathcal{C}}{2}\sum_{i=1}^n \mathcal{L}_{wave}(\xi_i) \nonumber \\
    &s.t.~~ z_i\beta=[x_i, h(x_i)]\beta - Y_i = \xi_i, ~\forall i \in \{1,2,\hdots,n\},
\end{align}
where $z_i=[x_i, h(x_i)]$, $h(x_i)$ is the hidden layer corresponding to the sample $x_i$,  $\mathcal{C}$ represents the tunable parameter, $\beta$ is the output layer matrix and \(\xi_i\) represents the error variable, allowing tolerance for misclassifications in situations of overlapping distributions. Putting the values of $\xi_i$ in the optimization function of \eqref{Eq:1}, we obtain
\begin{align}
    \min f(\beta) & = \min\frac{1}{2}\|\beta\|^2 + \frac{\mathcal{C}}{2}\sum_{i=1}^n \mathcal{L}_{wave}(z_i\beta - Y_i),
\end{align}
\vspace{-8mm}
where $\mathcal{L}_{wave}$ is the wave loss function defined in \eqref{Wave:1}. \\ 
Then, the optimization is reduced to the following form:
\begin{align}
\label{Eq:10}
    \min f(\beta) & = \min\frac{1}{2}\|\beta\|^2 + \frac{\mathcal{C}}{2\eta} \sum_{i=1}^n \left( 1 - \frac{1}{1+\eta(z_i\beta - Y_i)^2e^{\gamma(z_i\beta - Y_i)}}   \right).
\end{align}
Optimizing the Wave-RVFL model presents challenges due to the non-convexity of the loss function. Nevertheless, by leveraging the inherent smoothness of Wave-RVFL, gradient-based algorithms can be employed for model optimization.

\subsection{Soultion: Wave-RVFL}
Now, we employ the adaptive moment estimation (Adam) algorithm for solving the optimizing problem \eqref{Eq:10}. The \eqref{Eq:10} can be written as:
\begin{align}
    \min f(\beta) & = \min\frac{1}{2}\|\beta\|^2 + \frac{\mathcal{C}}{2\eta} \sum_{i=1}^n \left( 1 - \frac{1}{1+\eta \xi_i^2e^{\gamma \xi_i}}   \right),
\end{align}
where $\xi_i = z_i\beta - Y_i$.\\

At $t^{th}$ iteration (maximum number of iterations is denoted as ``$Itr$''), \( s \) samples are selected randomly and used to find the gradient and the following steps. Taking the gradient with respect to $\beta$, we obtain:
\begin{align}
\label{eq:13}
   \frac{\partial f(\beta)}{\partial \beta} = \beta + \frac{\mathcal{C}}{2}\sum_{i=1}^s\left( \frac{\xi_i z_i^T e^{\gamma\xi_i} (2+ \xi_i\gamma)}{(1+\eta\xi_i^2e^{\gamma\xi_i})^2} \right). 
\end{align}
Following \cite{adam2019no}, we construct the first moment vector ($g_t$) and second moment vector $(u_t)$ as follows:
\begin{align}
  & g_t = \lambda_1 g_{t-1} + (1-\lambda_1)\frac{\partial f(\beta_t)}{\partial \beta_t}. \label{eq:14} \\
  & u_t = \lambda_2 u_{t-1} + (1-\beta_2) \left(\frac{\partial f(\beta_t)}{\partial \beta_t} \right)^2. \label{eq:15}
\end{align}
Here, \( \lambda_1 \) and \( \lambda_2 \) represent the decay rates for the first and second-moment estimates, commonly set to $0.9$ and $0.999$, respectively. Next, we compute the bias-corrected first and second moment estimates using:
\begin{align}
\hat{g}_t = \frac{g_t}{(1-\lambda_1^t)}. \label{eq:16}\\
    \hat{u}_t  = \frac{u_t}{(1-\lambda_2^t)}. \label{eq:17} 
\end{align}
Finally, we update the parameter $\beta$ as follows:
\begin{align}
\label{eq:18}
    \beta_t = \beta_{t-1} - \alpha\frac{\hat{g}_t}{\sqrt{\hat{u}_t} + \epsilon}.
\end{align}
Here, \( \epsilon>0 \) is a small constant, and \( \alpha \) represents the learning rate.\\
\textbf{Note:} Algorithm \ref{Wave-RVFL classifier} outlines the details of calculating $\beta$ of the proposed Wave-RVFL model. The time complexity of the proposed model is $\mathcal{O}(Itr(nmN + s)),$ with detailed information provided in Supplementary Section S.II.
\begin{algorithm}
\caption{Wave-RVFL classifier}
\label{Wave-RVFL classifier}
\textbf{Input:} Let $\{x_i\}_{i=1}^n$ be the input training dataset, and $Y_i$ be the target output. $\mathcal{C}$, $\eta$, and $\gamma$ are the trade-off parameters. \\
\textbf{Output:} The Wave-RVFL parameter $\beta$.
\begin{algorithmic}[1]
\STATE Initialize: $\alpha_{0}$ and $t$.\\
\STATE Select \( s \) samples \( \{x_i\}_{i=1}^s \) through uniform random sampling.\\
\STATE Calculate $\frac{\partial f(\beta)}{\partial \beta}$ using \eqref{eq:13}.
\STATE Calculate $g_t$ using \eqref{eq:14}.
\STATE Calculate $u_t$ using \eqref{eq:15}.
\STATE Calculate $\hat{g}_t$ using \eqref{eq:16}.
\STATE Calculate $\hat{u}_t$ using \eqref{eq:17}.
\STATE Update the Wave-RVFL's output layer weight $\beta_t$ using \eqref{eq:18}.
\STATE Increment the iteration counter: \( t = t + 1 \).
\end{algorithmic}
\textbf{Until:} $\lvert \beta_t - \beta_{t-1}\rvert < \delta$ or $t=Itr$.\\
\textbf{Return:} $\beta_t$.
\end{algorithm}
\vspace{-8mm}
\section{Experiments, Results and Discussion}
\vspace{-3mm}
\label{experiments}
This section presents comprehensive details of the experimental setup, datasets, and compared models. Subsequently, we delve into the experimental results and conduct statistical analyses. We also examine the influence of noise and outliers on the performance of the proposed Wave-RVFL. At last, we conduct sensitivity analyses for various hyperparameters of the proposed Wave-RVFL.
\vspace{-4mm}
\subsection{Compared Models, Datasets and Experimental Setup}
\vspace{-2mm}
We conduct comparisons among the proposed Wave-RVFL; and several benchmarks, including RVFL \cite{pao1994learning}, ELM (also known as RVFL without direct link (RVFLwoDL)) \cite{huang2006extreme}, intuitionistic fuzzy RVFL (IF-RVFL) \cite{malik2022alzheimer}, graph embedded ELM with linear discriminant analysis (GEELM-LDA) \cite{iosifidis2015graph}, graph embedded ELM with local Fisher discriminant analysis (GEELM-LFDA) \cite{iosifidis2015graph}, minimum class variance based ELM (MCVELM) \cite{6542653} and Neuro-fuzzy RVFL (NF-RVFL) whose fuzzy layer centers are generated using random-means \cite{sajid2024neuro}. 

To evaluate the efficacy of the proposed Wave-RVFL models, we employ $23$ benchmark datasets from the UCI \cite{dua2017uci} repository from various domains and sizes. For example, the number of samples in the ``fertility'' dataset is $100$, and the number of samples in the ``connect\_4'' dataset is $67,557$. The experimental setup is discussed in Section S.III of the supplementary material.
\begin{table}[]
\centering
\caption{Testing accuracy, standard deviation and average rank of all the baseline models along with the proposed Wave-RVFL model.}
\label{UCI and KELL results}
\resizebox{\textwidth}{!}{%
\begin{tabular}{lcccccccc} \hline
Model $\rightarrow$ &
  RVFLwoDL \cite{huang2006extreme} &
  RVFL \cite{pao1994learning} &
  GEELM-LDA \cite{iosifidis2015graph} &
  GEELM-LFDA \cite{iosifidis2015graph} &
  MCVELM \cite{6542653} &
  IFRVFL \cite{malik2022alzheimer}&
  NF-RVFL \cite{sajid2024neuro}&
  Wave-RVFL$^{\dagger}$ \\ \hline
Dataset $\downarrow$ &
  ACC $\pm$ Std &
  ACC $\pm$ Std &
  ACC $\pm$ Std &
  ACC $\pm$ Std &
  ACC $\pm$ Std &
  ACC $\pm$ Std &
  ACC $\pm$ Std &
  ACC $\pm$ Std \\ \hline
acute\_inflammation &
  $100 \pm 0$ &
  $100 \pm 0$ &
  $100 \pm 0$ &
  $100 \pm 0$ &
  $100 \pm 0$ &
  $100 \pm 0$ &
  $100 \pm 0$ &
  $100 \pm 0$ \\
adult &
  $83.8725 \pm 0.1637$ &
  $84.0342 \pm 0.1941$ &
  $83.422 \pm 1.7111$ &
  $83.1784 \pm 2.8045$ &
  $83.9011 \pm 0.3574$ &
  $83.5438 \pm 1.7036$ &
  $83.2992 \pm 0.5117$ &
  $76.3544 \pm 0.4048$ \\
bank &
  $89.4934 \pm 0.3245$ &
  $89.4051 \pm 0.5132$ &
  $89.6042 \pm 0.6524$ &
  $46.4966 \pm 11.7491$ &
  $89.6703 \pm 0.5543$ &
  $89.1173 \pm 0.7047$ &
  $89.4269 \pm 0.6225$ &
  $88.5202 \pm 0.5503$ \\
blood &
  $76.9056 \pm 13.1203$ &
  $76.5065 \pm 14.5247$ &
  $67.1732 \pm 26.7687$ &
  $76.2398 \pm 14.985$ &
  $77.3065 \pm 13.2094$ &
  $77.4398 \pm 14.3189$ &
  $77.838 \pm 11.8577$ &
  $76.5065 \pm 14.7975$ \\
breast\_cancer &
  $70.1754 \pm 44.6249$ &
  $70.1754 \pm 44.6249$ &
  $89.8246 \pm 22.753$ &
  $84.5796 \pm 22.9315$ &
  $70.5263 \pm 44.4418$ &
  $71.9298 \pm 28.2614$ &
  $72.3351 \pm 21.0696$ &
  $92.6316 \pm 16.4763$ \\
breast\_cancer\_wisc\_prog &
  $80.3846 \pm 8.7784$ &
  $81.359 \pm 4.3621$ &
  $62.0897 \pm 4.3205$ &
  $62.8462 \pm 26.7954$ &
  $81.8718 \pm 5.2731$ &
  $78.359 \pm 7.4479$ &
  $83.359 \pm 4.4786$ &
  $79.3462 \pm 6.3219$ \\
congressional\_voting &
  $63.2184 \pm 2.2989$ &
  $63.6782 \pm 4.4219$ &
  $59.7701 \pm 7.6676$ &
  $54.2529 \pm 3.671$ &
  $63.4483 \pm 2.3556$ &
  $58.8506 \pm 5.7125$ &
  $63.908 \pm 4.1919$ &
  $62.5287 \pm 3.4097$ \\
conn\_bench\_sonar\_mines\_rocks &
  $60.5226 \pm 5.489$ &
  $62.079 \pm 9.017$ &
  $80.5343 \pm 24.2302$ &
  $73.6585 \pm 36.1107$ &
  $64.4251 \pm 7.4481$ &
  $54.8316 \pm 6.5215$ &
  $65.8885 \pm 16.4039$ &
  $94.6341 \pm 11.9984$ \\
connect\_4 &
  $75.4059 \pm 3.8174$ &
  $75.4518 \pm 3.7595$ &
  $75.4518 \pm 0.6034$ &
  $75.4281 \pm 2.1291$ &
  $75.4459 \pm 3.7962$ &
  $75.3407 \pm 1.3324$ &
  $75.4844 \pm 3.9187$ &
  $76.3911 \pm 3.7587$ \\
fertility &
  $91 \pm 8.2158$ &
  $91 \pm 6.5192$ &
  $69 \pm 32.8634$ &
  $68 \pm 12.0416$ &
  $91 \pm 6.5192$ &
  $92 \pm 6.7082$ &
  $91 \pm 7.4162$ &
  $90 \pm 7.9057$ \\
haberman\_survival &
  $73.8181 \pm 8.5202$ &
  $73.4902 \pm 8.4751$ &
  $55.5632 \pm 4.4362$ &
  $52.2898 \pm 5.5721$ &
  $73.4902 \pm 8.4751$ &
  $75.1348 \pm 6.8822$ &
  $75.1296 \pm 7.1747$ &
  $76.4738 \pm 9.2421$ \\
hepatitis &
  $82.5806 \pm 10.3528$ &
  $85.1613 \pm 10.6011$ &
  $85.1613 \pm 9.5693$ &
  $85.8065 \pm 7.4264$ &
  $85.1613 \pm 9.5693$ &
  $85.8065 \pm 6.6892$ &
  $85.8065 \pm 13.0236$ &
  $84.5161 \pm 8.3498$ \\
ilpd\_indian\_liver &
  $72.2149 \pm 4.2892$ &
  $71.5311 \pm 5.5959$ &
  $61.7521 \pm 8.3776$ &
  $61.4353 \pm 8.5934$ &
  $72.5508 \pm 5.9868$ &
  $72.7277 \pm 5.9611$ &
  $72.3932 \pm 5.6986$ &
  $73.2508 \pm 5.3362$ \\
magic &
  $78.5279 \pm 15.2244$ &
  $78.7487 \pm 15.4047$ &
  $77.4289 \pm 0.9713$ &
  $78.3579 \pm 2.8589$ &
  $78.7592 \pm 15.7559$ &
  $77.5289 \pm 16.2896$ &
  $76.3407 \pm 7.31$ &
  $95.1682 \pm 10.7894$ \\
molec\_biol\_promoter &
  $74.632 \pm 8.0443$ &
  $72.7706 \pm 7.6472$ &
  $71.8182 \pm 20.158$ &
  $77.4892 \pm 15.167$ &
  $72.684 \pm 5.8889$ &
  $78.3983 \pm 7.4359$ &
  $82.0346 \pm 5.3415$ &
  $92.381 \pm 7.0367$ \\
musk\_1 &
  $69.7675 \pm 7.941$ &
  $72.0614 \pm 2.3802$ &
  $72.4496 \pm 29.6161$ &
  $65.9518 \pm 18.3889$ &
  $70.1776 \pm 5.1064$ &
  $71.864 \pm 7.1741$ &
  $75.2149 \pm 6.5728$ &
  $96.6316 \pm 7.532$ \\
parkinsons &
  $80.5128 \pm 19.5781$ &
  $80.5128 \pm 17.7276$ &
  $84.1026 \pm 19.3075$ &
  $83.5897 \pm 16.3782$ &
  $83.5897 \pm 18.1853$ &
  $78.4615 \pm 7.3871$ &
  $83.0769 \pm 12.6398$ &
  $87.1795 \pm 12.4299$ \\
planning &
  $71.3814 \pm 8.8534$ &
  $71.3814 \pm 8.8534$ &
  $59.1592 \pm 23.3447$ &
  $61.1261 \pm 22.3356$ &
  $73.048 \pm 8.5221$ &
  $69.8048 \pm 9.2549$ &
  $73.018 \pm 8.381$ &
  $73.6036 \pm 11.0985$ \\
ringnorm &
  $51.5405 \pm 1.191$ &
  $51.5541 \pm 1.4357$ &
  $51.5541 \pm 7.0865$ &
  $51.5784 \pm 1.7619$ &
  $51.9459 \pm 1.4217$ &
  $51.473 \pm 1.2444$ &
  $51.6081 \pm 0.5721$ &
  $51.8649 \pm 1.437$ \\
spambase &
  $87.0472 \pm 4.8721$ &
  $88.546 \pm 4.5479$ &
  $88.9582 \pm 6.8046$ &
  $88.546 \pm 3.5979$ &
  $87.4382 \pm 5.3044$ &
  $85.0883 \pm 7.8931$ &
  $88.9582 \pm 4.6103$ &
  $99.3913 \pm 1.3611$ \\
spect &
  $66.7925 \pm 6.3425$ &
  $68.3019 \pm 5.2356$ &
  $63.3962 \pm 7.9604$ &
  $62.6415 \pm 9.8403$ &
  $67.9245 \pm 8.7487$ &
  $68.3019 \pm 9.6577$ &
  $69.0566 \pm 10.0373$ &
  $66.0377 \pm 9.5278$ \\
spectf &
  $79.7205 \pm 20.7374$ &
  $79.3431 \pm 20.8936$ &
  $80.0978 \pm 20.0899$ &
  $80.8526 \pm 18.722$ &
  $79.7205 \pm 20.7374$ &
  $79.3431 \pm 20.8936$ &
  $79.7205 \pm 20.7374$ &
  $81.9846 \pm 16.8489$ \\
statlog\_heart &
  $80 \pm 3.3127$ &
  $80.3704 \pm 2.8085$ &
  $73.7037 \pm 4.0147$ &
  $74.8148 \pm 1.6563$ &
  $82.5926 \pm 1.0143$ &
  $81.8519 \pm 4.2229$ &
  $81.4815 \pm 3.7037$ &
  $80.3704 \pm 1.0143$ \\ \hline
\textbf{Average (ACC $\pm$ Std)} &
  $76.5006 \pm 8.9605$ &
  $76.8462 \pm 8.6758$ &
  $74.0007 \pm 12.3177$ &
  $71.7026 \pm 11.5442$ &
  $77.2469 \pm 8.6379$ &
  $76.3999 \pm 7.9868$ &
  $78.1034 \pm 7.6641$ &
  $\textbf{82.4246} \pm \textbf{7.2881}$ \\ \hline
  \textbf{Average Rank} &
 $5.2826$&	$4.6304$	&$5.2609$&	$5.7609$&	$3.7391$	&$4.9348$&	$3.2391$&	$\textbf{3.1522}$ \\ \hline
 \multicolumn{9}{l}{Here, $\dagger$ indicates the proposed model. Boldface denotes the best-performed model.}\\
\end{tabular}%
}
\end{table}
\begin{table}[]
\centering
\caption{Win Tie Loss test for the results tabulated in Table \ref{UCI and KELL results}.}
\label{Win Tie Loss Test}
\resizebox{0.7\linewidth}{!}{%
\begin{tabular}{l|ccccccc} \hline
 &
  \multicolumn{1}{c}{RVFLwoDL \cite{huang2006extreme}} &
  \multicolumn{1}{c}{RVFL \cite{pao1994learning}} &
  \multicolumn{1}{c}{GEELM-LDA \cite{iosifidis2015graph}} &
  \multicolumn{1}{c}{GEELM-LFDA \cite{iosifidis2015graph}} &
  \multicolumn{1}{c}{MCVELM \cite{6542653}} &
  \multicolumn{1}{c}{IFRVFL \cite{malik2022alzheimer}} &
  \multicolumn{1}{c}{NF-RVFL \cite{sajid2024neuro}} \\ \hline
RVFL \cite{pao1994learning}      & [$12, 5, 6$]  &              &               &              &               &              &              \\
GEELM-LDA \cite{iosifidis2015graph}  & [$10, 1, 12$] & [$7, 4, 12$] &               &              &               &              &              \\
GEELM-LFDA \cite{iosifidis2015graph} & [$9, 1, 13$]  & [$7, 2, 14$] & [$9, 1, 13$]  &              &               &              &              \\
MCVELM \cite{6542653}     & [$18, 3, 2$]  & [$12, 4, 7$] & [$14, 2, 7$]  & [$15, 2, 6$] &               &              &              \\
IFRVFL \cite{malik2022alzheimer}     & [$10, 1, 12$] & [$8, 3, 12$] & [$12, 1, 10$] & [$13, 2, 8$] & [$9, 1, 13$]  &              &              \\
NF-RVFL \cite{sajid2024neuro}  & [$17, 3, 3$]  & [$19, 2, 2$] & [$14, 2, 7$]  & [$16, 2, 5$] & [$12, 3, 8$]  & [$15, 2, 6$] &              \\
Wave-RVFL$^{\dagger}$  & [$15, 1, 7$]  & [$13, 3, 7$] & [$19, 1, 3$]  & [$20, 1, 2$] & [$12, 1, 10$] & [$15, 1, 7$] & [$13, 1, 9$] \\ \hline
\end{tabular}%
}
\end{table}
\vspace{-4mm}
\subsection{Experimental Results on UCI Dataset}
\vspace{-2mm}
In this section, we thoroughly analyze and compare the performance of the proposed Wave-RVFL model against baseline models using various statistical metrics and tests, including accuracy, rank, and Friedman test, to provide a robust comparison. Table \ref{UCI and KELL results} presents the experimental findings regarding accuracy (ACC), standard deviation (Std), and ranks.\\
\textbf{\textit{Accuracy and standard deviation}}: Our analysis based on ACC reveals that the proposed Wave-RVFL model consistently outperforms several baseline models, including RVFLwoDL, RVFL, GEELM-LDA, GEELM-LFDA, MCVELM, IFRVFL, and NF-RVFL, across most datasets. From Table \ref{UCI and KELL results}, it's clear that our proposed Wave-RVFL model achieves the highest average ACC at $82.4246\%$. In contrast, the average ACC of the RVFLwoDL, RVFL, GEELM-LDA, GEELM-LFDA, MCVELM, IFRVFL, and NF-RVFL models are $76.5006\%$, $76.8462\%$, $74.0007\%$, $71.7026\%$, $77.2469\%$, $76.3999\%$, and $78.1034\%$, respectively. The proposed Wave-RVFL model demonstrates minimal standard deviation values compared to the baseline models, indicating a high level of prediction certainty. On the breast\_cancer dataset, our proposed Wave-RVFL achieves an ACC of $92.6316\%$, showcasing exceptional performance compared to the baseline models, with an increase of around $22\%$ compared to the RVFL model. On the conn\_bench\_sonar\_mines\_rocks dataset, Wave-RVFL outperforms the baseline models by securing the top position with an increase of $14.10\%$. Similar observations are recorded for the other datasets as well. \\
\textbf{\textit{Statistical rank}}: Sometimes, the model's average accuracy metric may be skewed by exceptional performance on a single dataset, masking weaker results across others. This could lead to a biased assessment of its overall performance. Hence, we utilize a ranking methodology to evaluate the comparative efficacy of the models under consideration. In this ranking scheme, classifiers are assigned ranks based on their performance, with better-performing models receiving lower ranks and those with poorer performance receiving higher ranks. To assess the performance of \( q \) models across \( P \) datasets, we denote \( r_j^i \) as the rank of the \( j^{th} \) model on the \( i^{th} \) dataset. $\mathscr{R}_j = \frac{1}{P}\sum_{i=1}^P r_j^i$ is the average rank of the $j^{th}$ model. From the last row of Table \ref{UCI and KELL results}, we find that the average rank of the proposed Wave-RVFL model along with the RVFLwoDL, RVFL, GEELM-LDA, GEELM-LFDA, MCVELM, IFRVFL, and NF-RVFL models are $3.1522$, $5.2826$, $4.6304$, $5.2609$, $ 5.7609$, $3.7391$, $4.9348$, and $ 3.2391$, respectively. The Wave-RVFL model achieves the lowest average rank, indicating superior performance compared to all other models and demonstrating its strong generalization ability.\\
 \textbf{\textit{Friedman test}}: Now, we perform the Friedman test \cite{demvsar2006statistical} to determine if there are statistically significant differences among the compared models. Under the null hypothesis, it is assumed that all models exhibit an equal average rank, suggesting that their performance levels are comparable. The Friedman statistic follows the chi-squared distribution $(\chi_F^2)$ with $(q - 1)$ degrees of freedom (d.o.f), and its computation involves: $\chi_F^2 = \frac{12P}{q(q+1)}\left[ \sum_j \mathscr{R}_j^2 - \frac{q(q+1)^2}{4} \right]$. The $F_F$ statistic is computed as: $F_F = \frac{(P-1)\chi_F^2}{P(q-1) - \chi_F^2}$, where the $F$-distribution possesses degrees of freedom $(q - 1)$ and $(P - 1) \times (q - 1)$. With $q=8$ and $P=23$ in our case, we calculate $\chi_F^2 = 26.7289$ and $F_F = 4.3795$. The critical value $F_F(7, 154) = 2.0695$ at a $5\%$ level of significance. As the calculated test statistic of $4.3795$ surpasses the critical value of $2.0695$, then the null hypothesis is rejected. This indicates a statistically significant difference among the models under comparison. \\
\textbf{\textit{Win Tie Loss test}}: Next, we utilize the pairwise Win Tie Loss (W-T-L) test to assess the performance of the proposed model compared to baseline models. Table \ref{Win Tie Loss Test} presents a comparative analysis of the proposed Wave-RVFL model alongside the baseline models. It delineates their performance regarding pairwise wins, ties, and losses across UCI datasets. In Table \ref{Win Tie Loss Test}, the entry $[x, y, z]$ indicates the frequency with which the model listed in the row wins $x$ times, ties $y$ times and loses $z$ times when compared to the model listed in the corresponding column. In Table \ref{Win Tie Loss Test}, the W-T-L outcomes of the models are evaluated pairwise. The proposed Wave-RVFL model has achieved $15$ wins (against RVFLwoDL), $13$ wins (against RVFL), $19$ wins (against GEELM-LDA), $20$ wins (against GEELM-LFDA), $12$ wins (against MCVELM), $15$ wins (against IFRVFL ), and $13$ wins (against NF-RVFL) out of $23$ datasets. Thus, this indicates that the proposed Wave-RVFL model achieves superior performance compared to the baseline models.

Considering the above discussion based on accuracy, rank, and statistical tests, we can conclude that the proposed Wave-RVFL model showcases superior and robust performance against the baseline models.
\begin{table}[]
\centering
\caption{Performance of the proposed Wave-RVFL along with the RVFL with varying levels of outliers and noise.}
\label{tab:noise}
\resizebox{9cm}{!}{%
\begin{tabular}{|lccc|ccc|}
\hline
\multicolumn{1}{|l|}{}        & \multicolumn{3}{c|}{Experiments with outliers}          & \multicolumn{3}{c|}{Experiments with noise}            \\ \hline
\multicolumn{1}{|l|}{Dataset} & \multicolumn{1}{c|}{Outliers}     & RVFL \cite{pao1994learning}    & Wave-RVFL$^{\dagger}$ & \multicolumn{1}{c|}{Noise}        & RVFL \cite{pao1994learning}    & Wave-RVFL$^{\dagger}$ \\ \hline
\multicolumn{1}{|l|}{blood}   & \multicolumn{1}{c|}{$5\%$}         & 45.685  & \textbf{73.1615}   & \multicolumn{1}{c|}{$5\%$}         & 44.868  & \textbf{68.3463}   \\
\multicolumn{1}{|l|}{}        & \multicolumn{1}{c|}{$10\%$}  & 50.5647      & \textbf{71.6886}   & \multicolumn{1}{c|}{$10\%$}      & 71.6716 & \textbf{76.2398}   \\
\multicolumn{1}{|l|}{}        & \multicolumn{1}{c|}{$15\%$}        & 51.2161 & \textbf{69.0112}   & \multicolumn{1}{c|}{$15\%$}      & 50.7714 & \textbf{76.2398}   \\
\multicolumn{1}{|l|}{}        & \multicolumn{1}{c|}{$20\%$}        & 52.9333 & \textbf{65.1302}   & \multicolumn{1}{c|}{$20\%$}       & 57.3477 & \textbf{69.7969}  \\ \cline{2-7} 
\multicolumn{1}{|l|}{}        & \multicolumn{1}{c|}{Average}   & 50.0998 & \textbf{69.7479}   & \multicolumn{1}{c|}{Average}  & 56.1647 & \textbf{72.6557}   \\ \hline
\multicolumn{1}{|l|}{magic}   & \multicolumn{1}{c|}{$5\%$}         & 54.3428 & \textbf{63.4437}   & \multicolumn{1}{c|}{$5\%$}        & \textbf{69.0799} & 65.7992   \\
\multicolumn{1}{|l|}{}        & \multicolumn{1}{c|}{$10\%$}         & 37.7077 & \textbf{70.2103}   & \multicolumn{1}{c|}{$10\%$}        & 66.6404 & \textbf{84.837}    \\
\multicolumn{1}{|l|}{}        & \multicolumn{1}{c|}{$15\%$}        & 60.3207 & \textbf{60.4048}   & \multicolumn{1}{c|}{$15\%$}     & 68.0389 & \textbf{70.0053}   \\
\multicolumn{1}{|l|}{}        & \multicolumn{1}{c|}{$20\%$}       & 50.857  & \textbf{59.2429}   & \multicolumn{1}{c|}{$20\%$}       & \textbf{65.5363} & 64.837    \\ \cline{2-7} 
\multicolumn{1}{|l|}{}        & \multicolumn{1}{c|}{Average}   & 50.8071 & \textbf{63.3254}   & \multicolumn{1}{c|}{Average} &  67.3239 & \textbf{71.3696}   \\ \hline
\multicolumn{2}{|c|}{Overall Average}                           & 50.4534 & \textbf{66.5367}   & \multicolumn{1}{l|}{}         & 61.7443 & \textbf{72.0127}   \\ \hline
\multicolumn{2}{|c|}{Average Rank}                                   & 2    & \textbf{1}      & \multicolumn{1}{l|}{}              & 1.75     & \textbf{1.25}       \\ \hline
\multicolumn{7}{l}{Boldface in each row denotes the best-performed model.}\\
\end{tabular}%
}
\end{table}
\vspace{-3mm}
\subsection{Robustness Evaluation of the Proposed Wave-RVFL Model on Datasets with Outliers}
\vspace{-2mm}
To assess the robustness of the proposed Wave-RVFL model against outliers, we introduce label noise into the blood and magic datasets at levels of $5\%$, $10\%$, $15\%$, and $20\%$ to create outliers. This allowed a detailed analysis of the model's performance under adverse conditions, with results presented in Table \ref{tab:noise}.
\begin{enumerate} 
\item \textbf{\textit{``blood'' Dataset}}: The proposed Wave-RVFL model consistently outperformed the baseline models across all outlier levels, from $0\%$ to $20\%$. It achieved the highest average accuracy for the blood dataset at $69.7479\%$, demonstrating remarkable robustness.
\item \textbf{\textit{``magic'' Dataset}}: The Wave-RVFL model led the performance table in $4$ out of $4$ scenarios for the magic dataset. It recorded the highest average accuracy of $63.3254\%$, showcasing its effectiveness.
\item \textbf{\textit{Overall Performance}}: The Wave-RVFL model achieved an overall average accuracy of $63.5367\%$, approximately $16\%$ higher than the RVFL model. It also attained the lowest overall average rank of $1$, indicating its superior performance to the RVFL model.
\end{enumerate}
These findings illustrate that the proposed Wave-RVFL model displays exceptional robustness in challenging conditions.
\vspace{-3mm}
\subsection{Robustness Evaluation of the Proposed Wave-RVFL Model on Datasets with Gaussian Noise}
\vspace{-2mm}
To evaluate the robustness of the proposed Wave-RVFL model against noise, we introduce Gaussian noise into the magic and blood datasets at levels of $5\%$, $10\%$, $15\%$, and $20\%$. This approach provided a thorough comparison of the model's performance under adverse conditions against the RVFL, as shown in Table \ref{tab:noise}.
\begin{enumerate}
    \item \textbf{\textit{``blood'' Dataset}}: The Wave-RVFL model demonstrated exceptional robustness, consistently outperforming the baseline models across all outlier levels from $0\%$ to $20\%$. It achieved the highest average accuracy of $72.6557\%$ for the blood dataset.
    \item \textbf{\textit{``magic'' Dataset}}: Similarly, the Wave-RVFL model showcased its effectiveness by recording the highest average accuracy of $71.3696\%$ under varying levels of label noise for the magic dataset.
    \item \textbf{\textit{Overall Performance}}: The Wave-RVFL model's overall average accuracy is $72.0127\%$, approximately $10\%$ higher than the RVFL model. Furthermore, it attains the lowest overall average rank of $1.25$, underscoring its superior performance among all tested models.
\end{enumerate}
These observations show that the proposed Wave-RVFL model displays a notable insensitivity to noise, showcasing their superior robustness compared to the baseline models in challenging scenarios.
\vspace{-7mm}
\begin{figure*}
\begin{minipage}{.32\linewidth}
\centering
\subfloat[$\gamma$ v/s ACC]{\includegraphics[scale=0.18]{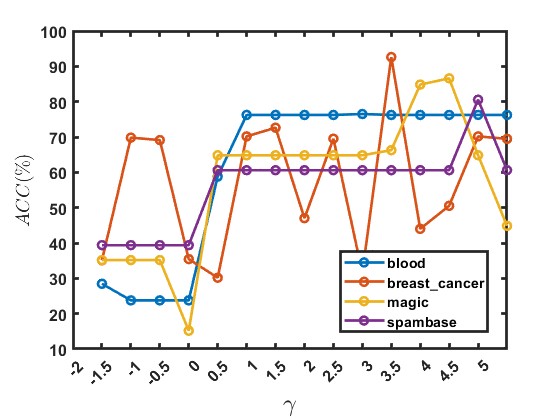}}
\end{minipage}
\begin{minipage}{.32\linewidth}
\centering
\subfloat[$\eta$ v/s ACC]{\includegraphics[scale=0.18]{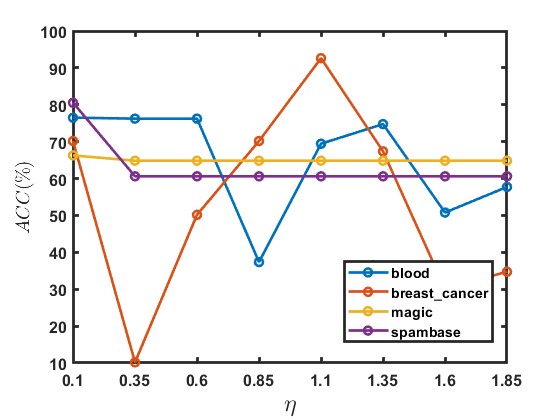}}
\end{minipage}
\begin{minipage}{.32\linewidth}
\centering
\subfloat[$\mathcal{C}$ v/s ACC]{\includegraphics[scale=0.18]{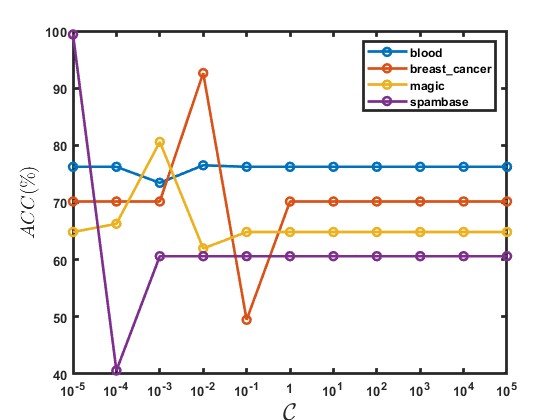}}
\end{minipage}
\par\medskip
\begin{minipage}{.32\linewidth}
\centering
\subfloat[$N$ v/s ACC]{\includegraphics[scale=0.18]{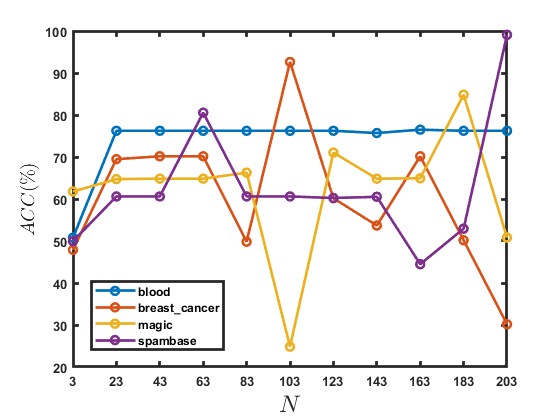}}
\end{minipage}
\begin{minipage}{.32\linewidth}
\centering
\subfloat[$Actfun.$ v/s ACC]{\includegraphics[scale=0.18]{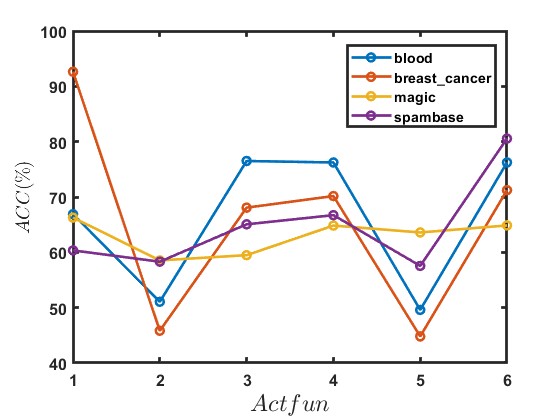}}
\end{minipage}
\begin{minipage}{.32\linewidth}
\centering
\subfloat[$\alpha$ v/s ACC]{\includegraphics[scale=0.18]{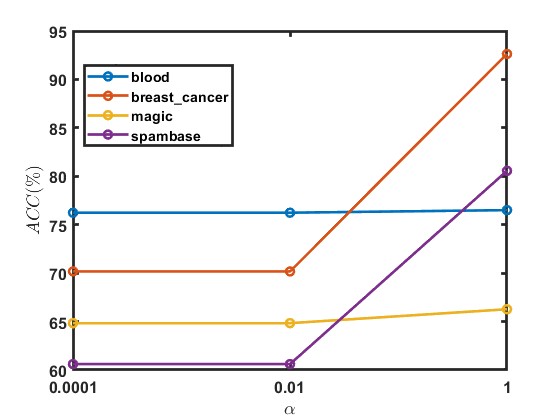}}
\end{minipage}
\caption{Effect of hyperparameters on the proposed Wave-RVFL model.}
\label{fig:effect_parameter}
\end{figure*}
\vspace{-8mm}
\vspace{-4mm}
\subsection{Sensitivity Analyses}
\vspace{-2mm}
To comprehensively understand the robustness of the proposed Wave-RVFL model, it is essential to analyze its sensitivity to various hyperparameters. Therefore, we conduct sensitivity analyses focusing on the following aspects: (i) $\gamma$ v/s ACC, (ii) $\eta$ v/s ACC, (iii) $\mathcal{C}$ v/s ACC, (iv) $N$ v/s ACC, (v) $Actfun$ v/s ACC, and (vi) $\alpha$ v/s ACC. We do experiment with varying ranges of each hyperparameter and assessed their influence on the model's performance using four datasets: blood, breast\_cancer, magic, and spambase. Insights from these analyses and from their corresponding Figure \ref{fig:effect_parameter} are detailed below.

\begin{enumerate}
    \item[(i)] \textbf{\textit{Effect of the wave loss hyperparameter $\gamma$}}: Analysis of Figure \ref{fig:effect_parameter}(a) indicates that increasing $\gamma$ generally enhances the model's performance up to a certain point. Optimal results are often observed when $\gamma$ is between $1$ and $4$. Thus, we recommend $\gamma$ to be varying in $[1, 4]$ for best performance.
    \item[(ii)] \textbf{\textit{Effect of the wave loss hyperparameter $\eta$}}: From Figure \ref{fig:effect_parameter}(b), it is evident that the model's performance improves as $\eta$ increases up to approximately $1.35$, beyond which performance declines. Therefore, for optimal results, $\eta$ should be set below $1.35$.
    \item[(iii)] \textbf{\textit{Effect of the regularization parameter $\mathcal{C}$}}: Examination of Figure \ref{fig:effect_parameter}(c) reveals that performance varies with $\mathcal{C}$ values, peaking around $\mathcal{C}=1$. Beyond this point, changes in $\mathcal{C}$ do not significantly impact performance. Consequently, we recommend $\mathcal{C}$ within the range of $10^{-5}$ to $1$ for optimal accuracy.
    \item[(iv)] \textbf{\textit{Effect of the hyperparameter $N$ (number of hidden nodes)}}: As observed in Figure \ref{fig:effect_parameter}(d), the impact of $N$ on performance is dataset-dependent. Therefore, it is advisable to fine-tune $N$ for each specific scenario.
    \item[(v)] \textbf{\textit{Effect of the hyperparameter $Actfun$ (activation function)}}: Figure \ref{fig:effect_parameter}(e) shows that certain activation functions, specifically $Sine$ (2) and $Tansig$ (5), consistently yield lower performance across datasets. Hence, it is prudent to exclude Sine and Tansig during activation function tuning.
    \item[(vi)] \textbf{\textit{Effect of the hyperparameter $\alpha$} (learning rate)}: According to Figure \ref{fig:effect_parameter}(f), the model's performance improves as the learning rate $\alpha$ increases. Therefore, setting $\alpha$ to $1$ is recommended for efficient performance.
\end{enumerate}
\vspace{-2mm}
It is important to note that the performance of Wave-RVFL may vary depending on the dataset and application (in line with the No Free Lunch Theorem \cite{adam2019no}). Therefore, tuning hyperparameters becomes crucial to improve the overall generalization performance of the proposed Wave-RVFL model.
\vspace{-3mm}
\section{Conclusion and Future Directions}
\vspace{-3mm}
\label{conclusions}
In this paper, we propose the Wave-RVFL model, which utilizes the wave loss function instead of the commonly used square loss function in RVFL. The wave loss function offers several advantages: it demonstrates robustness against noise and outliers by avoiding excessive penalization of deviations, thus achieving a balanced approach to handling noise and outliers. Additionally, the wave loss function enhances scalability by eliminating the need for matrix inversion when calculating the output layer weights. The proposed Wave-RVFL model's performance has been assessed using UCI benchmark datasets with sizes extending up to $67,557$ samples. Results demonstrate a notable improvement in average accuracy ranging from $4.5\%$ to $9\%$ compared to baseline models, with lower standard deviations observed across the board. This empirical evidence, supported by statistical tests, substantiates the superior performance of the proposed model relative to baseline models. Moreover, the robustness of the proposed Wave-RVFL model was assessed by introducing Gaussian noise and uniform outliers through label noise augmentation in the datasets. Once again, the results indicate the robustness of the proposed model compared to baseline models.

 In future work, we aim to compare our proposed models across different families of models comprehensively. The iterative nature of the Wave-RVFL model suggests the potential for the development of a closed-form solution loss function in future investigations. Moreover, we anticipate extending our proposed models to include deep and ensemble-based versions of RVFL in subsequent studies.
 \vspace{-2mm}
\section*{Acknowledgement}
\vspace{-2mm}
This study receives support from the Science and Engineering Research Board (SERB) through the Mathematical Research Impact-Centric Support (MATRICS) scheme Grant No. MTR/2021/000787. M. Sajid acknowledges the Council of Scientific and Industrial Research (CSIR), New Delhi, for providing fellowship under grants 09/1022(13847)/2022-EMR-I. 
\bibliography{refs.bib}
\bibliographystyle{abbrvnat}
\end{document}


\title{Supplementary Material of the Manuscript "Wave-RVFL: A Randomized Neural Network Based on Wave Loss Function"}
\author{M. Sajid \and A. Quadir \and
M. Tanveer}
\authorrunning{M. Sajid et al.}
%
\institute{Indian Institute of Technology Indore, Simrol, Indore, India 
\email{\{phd2101241003,mscphd2207141002,mtanveer\}@iiti.ac.in}\\
}
\maketitle              
%

\section{Square Loss Function}
\begin{figure}
    \centering
    \includegraphics[width=0.65\textwidth,height=5cm]{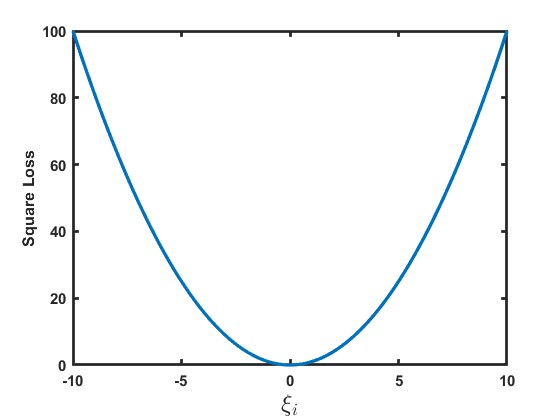}
    \caption{Visual depiction of squared loss function}
    \label{Squared loss}
\end{figure}

The square loss function \cite{wang2020comprehensive}, widely utilized in machine learning, calculates the error by squaring the difference between the actual value $y_i$ and the predicted value $f(x_i)$ for a given sample $x_i$, where $i=1, 2, \ldots, n$.
The square loss function is depicted in Figure \ref{Squared loss} and is defined as follows:
\begin{align}
    \mathcal{L}(y_i, f(x_i)) = \frac{1}{n}\sum_{i=1}^n \xi_i^2 = \frac{1}{n}\sum_{i=1}^n (y_i - f(x_i))^2,
\end{align}
where $\xi_i=y_i - f(x_i)$.
\\
The square loss function has the following drawbacks \cite{wang2020comprehensive}:
\begin{enumerate}
    \item[a)] The square error loss function gives higher weight to larger errors due to squaring. This means that outliers (data points that are far from the model's prediction) can disproportionately influence the loss, leading to suboptimal models.
    \item[b)] The square error loss function is scale-dependent, meaning it is sensitive to the magnitude of the errors. If the errors are large, the squared errors can become disproportionately large, impacting the training process and model performance.
    \item[c)] During optimization, the gradient of the squared error loss function can become very large for predictions that are far from the actual values. This can lead to unstable updates during gradient descent, causing the optimization process to become erratic or to diverge.
\end{enumerate}
\section{Computational Complexity}
The time complexity of the proposed Wave-RVFL model is influenced by two main components: (i) hidden layer generation and (ii) the application of the Adam algorithm for calculating $\beta$. According to \cite{sajid2024neuro}, the complexity of generating the hidden layer is $\mathcal{O}(nmN)$, where $n$ is the number of training samples, $m$ is the number of features, and $N$ represents the number of hidden layer nodes. In line with \cite{AKHTAR2024110637}, the Adam algorithm’s computational complexity primarily arises from computing the moving averages of the gradients’ first and second moments, along with bias correction for these averages. This step requires a complexity of $\mathcal{O}(s)$, where $s$ is the number of samples selected per iteration. Therefore, the total time complexity of the proposed Wave-RVFL model is $\mathcal{O}(Itr(nmN + s))$, where $Itr$ represents the number of iterations used in the Adam algorithm.
\section{Experimental Setup}
The experimental hardware configuration includes a personal computer featuring an Intel(R) Xeon(R) Gold 6226R CPU with a clock speed of 2.90 GHz and 128 GB of RAM. The system runs on Windows $11$ and utilizes Matlab2023a to run all the experiments. Following the experimental setup of \cite{sajid2024neuro}, we find the best hyperparamter setting and testing accuracy by employing grid search and five fold cross validation technique. Furthermore, all the hyperparameters of the baseline models are tune using the experimental setup followed in \cite{sajid2024neuro}.
For all the models (baseline and proposed), following \cite{sajid2024neuro}, we tune $6$ activation functions: $1$ denotes \textit{Sigmoid}, $2$ denotes \textit{Sine}, $3$ denotes \textit{Tribas}, $4$ denotes \textit{Radbas}, $5$ denotes \textit{Tansig} and $6$ denotes \textit{Relu}. The regularization parameter $\mathcal{C}$ is taken from $\{10^{-5}, 10^{-4}, \ldots, 10^5\}$. The number of hidden nodes ($N$) is selected from a range spanning from $3$ to $203$, with a step size of $20$. wave loss  parameters are selected from the following range $\eta = [0.1:0.25:2]$ and $\gamma = [-2 : 0.5 : 5]$. The Adam algorithm is initialized with the following parameters: starting weights \( \beta_0 = 0.01 \), initial learning rate \( \alpha \) selected from the set \(\{0.0001, 0.001, 0.01\}\), initial first moment \( g_0 = 0.01 \), initial second moment \( u_0 = 0.01 \), first-order exponential decay rate \( \lambda_1 = 0.9 \), second-order exponential decay rate \( \lambda_2 = 0.999 \), error tolerance \( \delta = 10^{-5} \), division constant \( \epsilon = 10^{-8} \), maximum number of iteration \( Itr = 1000 \), and mini-batch size \( s = 2^5 \) if number of samples in the dataset is less than $500$, otherwise $2^8$.
\bibliography{refs.bib}
\bibliographystyle{abbrvnat}